\def\year{2019}\relax
\documentclass[letterpaper]{article} %DO NOT CHANGE THIS
\usepackage{aaai19}  %Required
\usepackage{times}  %Required
\usepackage{helvet}  %Required
\usepackage{courier}  %Required
\usepackage{url}  %Required
\usepackage{graphicx}  %Required
\usepackage{times}  %Required
\usepackage{helvet}  %Required
\usepackage{courier}  %Required
\usepackage{url}  %Required
\usepackage{graphicx}  %Required
\usepackage{array}
\usepackage{booktabs} %调整表格线与上下内容的间隔
\usepackage{multirow}

\usepackage{amssymb,amsmath}
\usepackage{epsfig,epstopdf}
\usepackage{subfigure}
\usepackage{multirow}
\usepackage{float}
\usepackage[ruled]{algorithm}
\usepackage{algorithmic}
\usepackage{comment}
\usepackage{placeins}
\usepackage{microtype}
\usepackage{balance}
\usepackage{color}
\usepackage[normalem]{ulem}
\usepackage{bm}
\newcommand{\Rmnum}[1]{\expandafter\@slowromancap\romannumeral #1@}
\newcommand{\tabincell}[2]{
\begin{tabular}{@{}#1@{}}#2\end{tabular}
}
\begin{document}
\title{Ranking-based Deep  Cross-modal Hashing}
\author{Xuanwu Liu$^1$, Guoxian Yu$^{1,2}$\thanks{Corresponding author: gxyu@swu.edu.cn(Guoxian Yu)}, Carlotta Domeniconi$^3$, Jun Wang$^1$, Yazhou Ren$^4$, Maozu Guo$^5$\\
$^1$College of Computer and Information Sciences, Southwest University, Chongqing, China\\
$^2$Hubei Key Laboratory of Intelligent Geo-Information Processing, China University of Geosciences, Wuhan, China\\
$^3$Department of Computer Science, George Mason University, Fairfax, USA\\
$^4$SMILE Lab, School of Computer Science and Engineering, University of Electronic Science and Technology of China, Chengdu, China\\
$^5$School of Electrical and Information Engineering, Beijing University of Civil Engineering and Architecture, Beijing, China\\
Email: \{alxw1007,gxyu,kingjun\}@swu.edu.cn, carlotta@cs.gmu.edu, yazhou.ren@uestc.edu.cn, guomaozu@bucea.edu.cn\\
}

%\author{AAAI Press\\
%Association for the Advancement of Artificial Intelligence\\
%2275 East Bayshore Road, Suite 160\\
%Palo Alto, California 94303\\
%}
\maketitle
\begin{abstract}
Cross-modal hashing  has been receiving increasing interests for its low storage cost and fast query speed in multi-modal data retrievals. However, most existing hashing methods are based on hand-crafted  or raw level features of objects,  which may not be optimally compatible with the
coding process. Besides, these hashing methods are mainly designed
to handle simple pairwise similarity. The complex multilevel ranking
semantic structure of instances associated with multiple labels
has not been well explored yet. In this paper, we  propose a ranking-based deep  cross-modal hashing approach (RDCMH). RDCMH firstly uses the feature and label information
of data to derive a semi-supervised semantic ranking list. Next, to expand the semantic representation
power of hand-crafted features, RDCMH integrates the semantic ranking information into deep
cross-modal hashing and jointly optimizes the compatible parameters
of deep feature representations and of hashing functions. Experiments
on real multi-modal datasets show
that RDCMH outperforms other competitive baselines and achieves
the state-of-the-art performance in cross-modal retrieval
applications.
\end{abstract}

\section{Introduction}
With the explosive growth of data, how to efficiently and accurately retrieve the required information from massive data becomes a hot research topic and has various applications. For example, in information retrieval, approximate nearest neighbor (ANN) search \cite{Andoni2006Near} plays a fundamental role. Hashing has received increasing attention due to its low storage cost and fast retrieval speed  for  ANN search \cite{Kulis2010Kernelized}.  The main idea of hashing is to convert the high-dimensional data in the ambient  space into  binary codes in the low-dimensional Hamming space, while the proximity between data in the original space is preserved in the Hamming space\cite{wang2016L2H,wang2018L2H,Shao2016online}.
By using binary hash codes to represent the original data, the storage cost can
be dramatically reduced. In addition, we can use hash codes to construct an index and achieve a
constant or sub-linear time complexity for ANN search. Hence, hashing has become more and more popular for ANN search on large scale
datasets.

In many applications, the data can have multi-modalities. For
example, a web page can include not only a textual description but also images and videos to illustrate its contents. These different
types (views) of data are  called \emph{multi-modal} data.
With the rapid growth of multi-modal data in various applications,  multi-modal hashing has recently been widely
studied. Existing multi-modal hashing methods can be divided into two main
categories: mutli-source hashing (MSH)
and cross-modal hashing (CMH)\cite{zhu2013linear}. The
goal of MSH is to learn hash codes by utilizing all the information
from multiple modalities. Hence, MSH requires
 all the modalities  observed for all data points,
including query points and those in the database. In practice,  it
is often difficult or even infeasible to acquire all data points across all the modalities, as such
the application of MSH is limited. On
the contrary, the application scenarios of CMH are more
flexible and practical. In CMH, the modality of a
query point can be different from the modality of the points in the
database. In addition, the query point typically has only
one modality and the points in the database can have one or
more modalities. For example, we can use text queries to
retrieve images from the database, and we can also use image
queries to retrieve texts from the database. Due to its wide
application, CMH has attracted increasing attention \cite{Kumar2011Learning,Zhang2014Large}.

Many CMH methods have  been proposed recently, existing CMH methods can be roughly divided into two categories: supervised and unsupervised. Unsupervised approaches seek hash coding functions by taking into account underlying data structures, distributions, or topological information.
To name a few, Canonical correlation analysis \cite{rasiwasia2010new} maps two modalities, such as visual and textual, into a common space by
maximizing the correlation between the projections of the two modalities. Inter-media hashing \cite{Song2013Inter} maps view-specific features onto a common Hamming space by learning linear hash functions with intra-modal and inter-modal consistencies. Supervised approaches  try to leverage supervised information (i.e., semantic labels)
to improve the performance. Cross-modal similarity sensitive hashing (CMSSH) \cite{Bronstein2010Data} regards the  hash codes learning as binary classification problems, and efficiently learns the hash functions using a boosting method. Co-regularized hashing \cite{Yi2012Co} learns a group of hash functions for each bit of binary codes in every modal. Semantic correlation maximization (SCM) \cite{Zhang2014Large} optimizes the hashing functions by maximizing the correlation between
two modalities with respect to the semantic labels. Semantics Preserving Hashing
(SePH)\cite{Lin2017Cross} generates one unified hash code
for all observed views of any instance by considering the
semantic consistency between views.

Most supervised hashing are pairwise supervised methods, which leverage labels of instances and
pairwise labels of instance-pairs to train the coding functions,
such that the label information can be preserved in the
Hamming space \cite{Chang2012Supervised}. Their objectives, however, may
be suboptimal for ANN search, because they do not fully explore the high-order ranking information \cite{Song2015Top}. For example, a triplet rank contains a query
image, a positive image, and a negative image, where the
positive image is more similar to the query image than the
negative image \cite{Lai2015Simultaneous}. High-order ranking information carries relative similarity
ordering in the triplets and provides richer supervision, it often can be more easily obtained than pairwise ranking. Some hashing methods consider the high-order ranking information for hashing learning.  For example, deep semantic ranking based hashing \cite{Zhao2015Deep} learns deep hash functions based on CNN (Convolutional neural network)\cite{krizhevsky2012cnn}, which preserves the semantic structure of multi-label images. Simultaneous feature learning and hash coding \cite{Lai2015Simultaneous} generates bitwise hash codes for images via a carefully designed deep architecture and uses a triplet ranking loss to preserve relative similarities.

However, these semantic ranking methods just consider one modality, and cannot apply to cross-modal retrieval. Besides, the ranking lists are just simply computed by the number of shared labels, which could not preserve the integral ranking information of labels. Furthermore, the ranking lists adopted by these methods ask for sufficient labeled training data, and cannot make use of abundant unlabeled data, whose multi-modal feature information can boost the cross-modal hashing performance. Semi-supervised hashing methods were introduced to leverage both labeled and unlabeled samples \cite{Wang2012Semi}, but these methods cannot directly be applied on multi-modal data. Almost all these CMH
methods are based on hand-crafted (or raw-level) features. One
drawback of these hand-crafted feature based methods
is that the feature extraction procedure is isolated from
the hash-code learning procedure, or the original raw-level features can not reflect the semantic similarity between objects very well. The
hand-crafted features might not be optimally compatible
with the hash-code learning procedure\cite{Cao2017collective}. As a result, these CMH methods can not achieve satisfactory performance in real applications.

Recently, deep learning has also been utilized to perform feature learning from
scratches with promising performance.  Deep cross-modal hashing(DCMH)\cite{Jiang2017Deep} combines the deep feature learning with  cross-modal retrieval and guides deep learning procedure with multi-labels of multi-modal objects. Correlation auto-encoder hashing \cite{Cao2016Correlation} adopts deep learning for uni-modal
hashing. Their studies show that the end-to-end deep learning architecture is more compatible for hashing learning. However, they still ask for sufficient label information of training data, and treat the parameters of the hash quantization layer and those of deep feature learning layers as the same, which may reduce the discriminative power of the quantification process.

In this paper, we propose a  ranking-based deep cross-modal hashing (RDCMH), for cross-modal retrieval
applications. RDCMH firstly uses the feature and label information of data to derive a semi-supervised semantic ranking list. Next, it integrates the semantic ranking information into deep cross-modal hashing and jointly optimizes the ranking loss and hashing codes functions to seek optimal parameters of deep feature representations and those of hashing functions. The main contributions of RDCMH are outlined as follows:
\begin{enumerate}
\item A novel cross-modal hash function learning framework (RDCMH) is proposed to integrate deep feature learning with semantic
ranking   to address the problem
of preserving  semantic similarity between multi-label
objects for cross-modal hashing; and a label and feature information induced semi-supervised semantic ranking metric is also introduced to leverage labeled and unlabeled data.

\item RDCMH jointly optimizes the deep feature extraction process and the hash quantization process to make feature learning procedure being more compatible with the hash-code learning procedure, and this joint optimization indeed significantly improves the performance.

\item Experiments on benchmark multi-modal datasets
show that RDCMH outperforms other baselines \cite{Bronstein2010Data,Zhang2014Large,Lin2017Cross,Jiang2017Deep,Cao2016Correlation} and
 achieves the state-of-the-art performance in cross-modal
retrieval tasks.
\end{enumerate}

\section{The Proposed Approach}
Suppose $\mathbf{X}={\{x_1, x_2, \cdots, x_n}\} \in \mathbb{R}^{n \times d_X}$ and $\mathbf{Y}={\{y_1, y_2, \cdots, y_n}\} \in \mathbb{R}^{n \times d_Y}$ are two data modalities, $n$ is the number of instances (data points), $d_X$($d_Y$) is the dimensionality of the instances in the respective modality. For example, in the Wiki-image search application, $x_i$ is the image features of the entity $i$, and $y_i$  is the tag features of this entity. $\mathbf{Z} \in \mathbb{R}^{n \times m}$ stores the label information of $n$ instances in $\mathbf{X}$ and $\mathbf{Y}$ with respect to $m$ distinct labels. $z_{ik}\in{\{0,1}\}$, $z_{ik}=1$ indicates that $x_i$ is labeled with the $k$-th label; $z_{ik}=0$ otherwise. Without loss of generality, suppose the first $l$ samples have known labels, whereas other $u=n-l$ samples lack label information.  To enable cross-modal hashing, we need to learn two hashing functions, $F_1$: $\mathbb{R}^{d_{1}}\rightarrow {\{0,1}\}^{c}$ and $F_2$: $\mathbb{R}^{d_{2}}\rightarrow {\{0,1}\}^{c}$, where $c$ is the length of binary hash codes. These two hashing functions are expected to map the feature vectors in the respective modality onto a common Hamming space and to preserve the proximity of the original data.

RDCMH mainly involves with two steps. It firstly measures the semantic ranking between instances based on the label and feature information. Next, it defines an objective function to simultaneously account for semantic ranking, deep feature learning and hashing coding functions learning; and further introduces an alternative optimization procedure to jointly optimize these learning objectives. The overall workflow of RDCMH is shown in Fig. \ref{Fig1}.

%to obtain high-quality features of respective modality for hashing functions learning, and to ensure the obtained features being optimally compatible with triplet ranking loss and quantitative loss,
\begin{figure*}[h!t]
\centering
{\label{Fig1}\includegraphics[width=18cm,height=5.9cm]{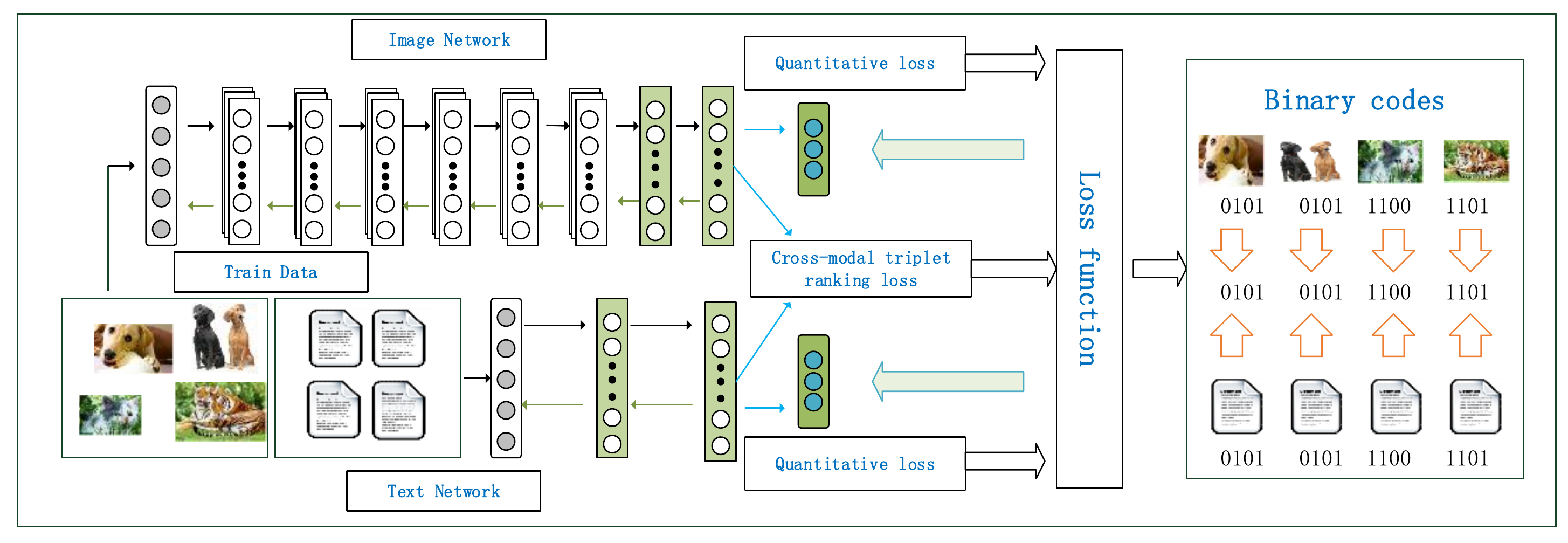}}
\caption{Workflow of the proposed  Rank based Deep Cross-Modal Hashing (RDCMH). RDCMH encompasses two steps: (1) an Image CNN network for learning image representations and a Text two-layer Network for learning text representations. (2) Jointly optimize the cross-modal triplet ranking loss and the quantitative loss to seek optimal parameters of deep feature representations and those of hashing functions.}
\label{Fig1}
\end{figure*}

\subsection{Semi-supervised Semantic Ranking }
To preserve the semantic structure, we can force the ranking order of neighbors computed by the Hamming distance being consistent with that derived from semantic labels in terms of ranking evaluation measures. Suppose $q$ is a query point, the semantic similarity level of a database point $x$ with respect to $q$ can be calculated based on ranking order of label information. Then we can obtain a ground-truth ranking list for $q$ by sorting the database points in decreasing order of their similarity levels \cite{Zhao2015Deep,Song2015Top}. However, this similarity level is just simply derived from the number of shared  labels  and these semantic ranking-based methods ignore that the labels of training data are not always readily available. Furthermore, these methods work on one modality and can not directly apply on multi-modal data.

To alleviate the issue of insufficient labeled data, we introduce a semi-supervised semantic measure that takes into account both the label and feature information of training data. The labels of an instance depend on the features of this instance, and the semantic similarity is positively correlated with the feature similarity of respective instances \cite{zhang2010multilabel,wang2009multi}. The semi-supervised semantic measure is defined as follows:
\begin{equation}
s_{ij}^{xx}=\left\{
             \begin{array}{lr}
            s_{ij}^1e^{(s_{ij}^2-s_{ij}^1)}, & |z_i|^2\neq {0}\ \ \textrm{and} \ \ |z_j|^2\neq {0}  \\
             s_{ij}^1, & \textrm{otherwise}
             \end{array}
\right.
\label{eq1}
\end{equation}
where $s_{ij}^1$ is the feature similarity of $x_i$ and $x_j$, while $s_{ij}^2$ is the label similarity, both of them are computed by the cosine similarity. Note, $s_{ij}^{xx}$ is always in the interval [0,1] and other similarity metrics can also be used. Eq. (\ref{eq1}) can account for both the labeled and unlabeled training data. Specifically, for two unlabeled data, the similarity between  $x_i$ and $x_j$ is directly computed from the feature information of the respective data. For labeled data, we consider that the label similarity  $s_{ij}^2$ is a supplement to $s_{ij}^1$. The larger the $s_{ij}^2$ is, the larger the $s_{ij}^{xx}$ is. In this way, we leverage the label and feature information of training data to account for insufficient labels.

Extending the ranking order to the cross-modal case, we should keep the sematic structure both in the inter-modality and intra-modality. Based on $\mathbf{S}^{xx} \in \mathbb{R}^{n \times n}$,  we  can obtain a ranking list $\{x_{q_k}\}_{k=1}^n$ for $q$ by sorting the database points
in decreasing order of $s^{xx}_{q_k}$. Similarly, we can define the semi-supervised
semantic similarity $\mathbf{S}^{yy} \in \mathbb{R}^{n \times n}$ for the data modality $\mathbf{Y}$. To balance the inconsistence of ranking list between two modalities, the semi-supervised semantic similarity is averaged as:  $\mathbf{S}^{xy}=\mathbf{S}^{yx}=(\mathbf{S}^{xx}+\mathbf{S}^{yy})/2$. Finally, we can obtain three different ranking lists: ${\{r^x_i}\}_{i=1}^n$, ${\{r^y_i}\}_{i=1}^n$, ${\{r^{xy}_i}\}_{i=1}^n$ for each query point.

\subsection{Unified Objective Function}
\subsubsection{Deep Feature Representation}\
Most existing hashing methods first extract hand-crafted visual features
(like GIST and SIFT) from images and then learn `shallow'
(usually linear) hash functions upon these features \cite{Bronstein2010Data,Zhang2014Large,Lin2017Cross}. However,
these hand-crafted features have limited representation
power and may lose key semantic information, which
is important for similarity search. Here we consider
designing deep hash functions using CNN \cite{krizhevsky2012cnn} to jointly
learn feature representations and
their mappings to hash codes. This non-linear hierarchical
hash function has more powerful learning capability than
the shallow one based on features crafted in advance, and
thus is able to learn more suitable feature representations
for multilevel semantic similarity search. Other representation learning models (i.e., AlexNet) can also be used  to learn deep features of images and text for RDCMH.
The feature learning part contains two deep neural networks, one for image modality and the other for text modality.

The adopted deep neural network for image modality is a CNN, which includes eight layers. The first six layers
are the same as those in CNN-F\cite{Chatfield2014Return}. The seventh and eighth layer
is a fully-connected layer with the outputs being the learned
image features. As to the text modality, we first represent
each text as a vector with bag-of-words (BOW) representation.
Next, the bag-of-words vectors are used as
the inputs for a neural network with two fully-connected
layers, denoted as ``full1 - full2''. The ``full1'' layer has 4096 neurons, and the second layer ``full2'' has $c$ (hashing codes) neurons, The activation function for the first layer is ReLU,
and that for the second layer is the identity function.

For presentation, we represent the learnt deep feature representations of $x$ and $y$  as $\varphi(x)$ and $\phi(y)$. The non-linear mapping parameters of these two representations will be discussed later.

\subsubsection{Triplet Ranking Loss and Quantitative Loss}\
Directly optimizing the ranking criteria for cross-modal hashing is very hard. Because it is very difficult to compare the ranking lists and stringently comply with the lists. To circumvent this problem, we use a triplet ranking loss  as the surrogate loss. Given a query $q$ and a ranking list $\{r^x_{q_i}\}_{i=1}^n$ for $q$, we can define a ranking loss on a set of triplets of hash codes as follows:
\begin{equation}
L(h(q),h({\{r^x_{q_i}\}}_{i=1}^n))\!=\!\sum_{i=1}^n\!\sum_{j: s_{qi}>s_{qj}}\![\delta dH(h(q),h(i),h(j))]_+
\label{eq2}
\end{equation}
where $n$ is the length of the ranking list, $s_{qi}$ and $s_{qj}$ are the similarity between query $q$ and  $x_i$ and $x_j$, respectively. $h(x)$ represents the learnt hash codes of $x$, $[x]_+=max(0,x)$, $\delta dH(a_1,a_2,a_3)=dH(a_1,a_2)-dH(a_1,a_3)$, $dH(\cdot)$ is the Hamming distance. This triplet ranking loss is a convex upper bound on the
pairwise disagreement, it counts the number of incorrectly ranked triplets.

Eq. (\ref{eq2}) equally treats all triplets, but two samples ($x_i$ and $x_j$) of a triplet may have different similarity levels to the query $q$. So we introduce the weighted ranking triplet loss based on the ranking list as follows:
\begin{equation}
\begin{split}
L(h(q),h({\{r^x_{q_i}\}}_{i=1}^n))=\sum\nolimits_{i=1}^n\sum\nolimits_{j:s_{qi}>s_{qj}}\\(1-s^{xx}_{ij})[\delta dH(h(q),h(i),h(j))]_+
\label{eq3}
\end{split}
\end{equation}
The larger the relevance between $x_i$ and $q$ than that between $x_j$ and $q$ is, the larger the ranking loss results in, if $x_i$ is ranked behind $x_j$ for $q$.

As to the cross-modal case, we should balance the inconsistence of ranking lists between two modalities. %{\color{blue}Specifically, for  a query $q$, we should simultaneous calculate $L(h(q),h({\{r^x_{q_i}\}}_{i=1}^n)),L(h(q),h({\{r^y_{q_i}\}}_{i=1}^n)),$
%$L(h(q),h({\{r^{xy}_{q_i}\}}_{i=1}^n))$, and $L(h(q),h({\{r^{yx}_{q_i}\}}_{i=1}^n))$.}
To this end, we give the unified objective function that simultaneously account for triplet ranking loss and quantitative loss as follows:
\begin{equation}
\begin{split}
\mathop {\min }\limits_{{\textbf{W}_x},{\textbf{W}_y},{\textbf{B}}} L&=\sum\nolimits_{q\in \mathcal{Q}}\sum\nolimits_{i,j=1}^n [(1-s_{ij}^{xx})({\delta dH_{xx}})+(1-s_{ij}^{yy}) \\&({\delta dH_{yy}})+(1-s_{ij}^{yx}) ({\delta dH_{yx}})+(1-s_{ij}^{xy}) ({\delta dH_{xy}})]\\&+{\frac{\lambda}{2}}(||\mathbf{B}^{x}-\mathbf{F}||^2_F+||\mathbf{B}^{y}-\mathbf{G}||^2_F)
\end{split}
\label{eq4}
\end{equation}
where
\begin{equation}
\begin{split}
&\delta dH_{xx}=dH(h(x_q),h(x_i))-dH(h(x_q),h(x_j))\\
&\delta dH_{yy}=dH(h(y_q),h(y_i))-dH(h(y_q),h(y_j))\\
&\delta dH_{xy}=dH(h(x_q),h(x_i))-dH(h(y_q),h(y_j))\\
&\delta dH_{yx}=dH(h(y_q),h(y_i))-dH(h(x_q),h(x_j))
\end{split}
\label{eq5}
\end{equation}

\begin{equation}
\begin{split}
h(x)=h(\varphi(x);\mathbf{W}_x)=sign(\mathbf{W}^T_x\varphi(x))\\
h(y)=h(\phi(y); \mathbf{W}_y)=sign(\mathbf{W}^T_y\phi(y))
\end{split}
\label{eq6}
\end{equation}

\begin{equation}
\begin{split}
\mathbf{F}_{*i}=(\varphi(x_i);\mathbf{W}_x), \
\mathbf{G}_{*i}=(\phi(y_i);\mathbf{W}_y) \\
\mathbf{B}^{x}\in {\{-1,+1}\}^{n\times c}, \
\mathbf{B}^{y}\in {\{-1,+1}\}^{n\times c}
\end{split}
\label{eq7}
\end{equation}
$\mathcal{Q}$ is the set of query points, $\varphi(x)$ and $\phi(y)$ are the  deep features of images and texts, $\mathbf{W}_x$ and $\mathbf{W}_y$ are the coefficient matrices of two modalities, respectively. $\lambda$ is the scalar parameter to balance the triplet ranking loss and quantitative loss.  $\mathbf{B}^{x}$  and $\mathbf{B}^{y}$ are the binary hash codes for image and text modality, respectively. In the training process,  since different modality data of the same sample share the same label set, and they actually represent the same sample from different viewpoints, we fix the binary codes of same training points from two modalities as the same, namely $\mathbf{B}^{x}=\mathbf{B}^{y}=\mathbf{B}$.

Eq. (\ref{eq4}) simultaneously accounts for the triplet ranking loss and the quantitative loss. The first term enforces the consistency of cross-modal ranking list by minimizing  the number of incorrectly ranked triplets, and the second term (weighted by $\lambda$) measures the quantitative loss of hashing.
$\mathbf{F}$ and $\mathbf{G}$ can preserve the cross-modal similarity in $\mathbf{S}^{xx}$, $\mathbf{S}^{yy}$ and $\mathbf{S}^{xy}$, as a result,
binary hash codes $\mathbf{B}^x$ and $\mathbf{B}^y$ can also preserve these cross-modal similarities. This exactly
coincides with the goal of cross-modal hashing.%GIVEN more analysis of the unified objective function. This 2nd main contribution of this paper.

%We exchangedly using $s_ij$, $r_ij$, they are different, ranking order is from 1 to n, but the similarity is between 0-1. We should avoid misleading the reviewer.}

\subsubsection{Optimization}
We can solve Eq. (\ref{eq4}) via the Alternating Direction Method of Multipliers (ADMM) \cite{Boyd2011Distributed},
which alternatively optimizes one of $\textbf{W}_x$, $\textbf{W}_y$, and $\textbf{B}$, while keeping the other two fixed.

\textbf{Optimize $\mathbf{W}_x$ with $\mathbf{W}_y$ and $\mathbf{B}$ fixed}:
We observe that the loss function in Eq. (\ref{eq4})
is actually a summation of weighted triplet
losses and the quantitative loss. Like most existing deep learning methods, we
utilize stochastic gradient descent (SGD) to learn $\mathbf{W}_x$ with
the back-propagation (BP) algorithm. In order to facilitate the gradient computation, we rewrite the
Hamming distance as the form of inner product: $dH(h(a),h(b))=\frac {c-h(a)^Th(b)}{2}$, where $c$ is the number of hash bits.

More specifically, in each iteration we
sample a mini-batch of points from the training set and then
carry out our learning algorithm based on the triplet data. For any  triplet ($q; x_i; x_j$), the derivative of Eq. (\ref{eq4}) with respect to coefficient matrix $\mathbf{W}_x$ in the data modality $\mathbf{X}$  is given by:
\begin{equation}
\begin{split}
\frac{\partial L}{\partial \mathbf{F}_{*q}} = \frac{1}{2}[(1-s^{xx}_{ij})(h(\mathbf{F}_{*i})-h(\mathbf{F}_{*j}))+(1-s^{xy}_{ij})\\(h(\mathbf{F}_{*i}))+(1-s^{yx}_{ij})(-h(\mathbf{F}_{*j}))]+\lambda(\mathbf{F}_{*q}-\mathbf{B}_{*q}) \\
\end{split}
\label{eq9}
\end{equation}

\begin{equation}
\begin{split}
\frac{\partial L}{\partial \mathbf{F}_{*i}} &= -\frac{1}{2}[(1-s^{xx}_{ij})(h(\mathbf{F}_{*q}))\\&+(1-s^{xy}_{ij})(h(\mathbf{F}_{*q}))]+\lambda(\mathbf{F}_{*i}-\mathbf{B}_{*i}) \\
\end{split}
\label{eq10}
\end{equation}

\begin{equation}
\begin{split}
\frac{\partial L}{\partial \mathbf{F}_{*j}} &= \frac{1}{2}[(1-s^{xx}_{ij})(h(\mathbf{F}_{*q}))\\&+(1-s^{xy}_{ij})(h(\mathbf{F}_{*q}))]+\lambda(\mathbf{F}_{*j}-\mathbf{B}_{*j})
\label{eq11}
\end{split}
\end{equation}
We can compute $\frac{\partial L}{\partial \mathbf{W}_{x}}$ with $\frac{\partial L}{\partial \mathbf{F}_{*q}}$, $\frac{\partial L}{\partial \mathbf{F}_{*i}}$ and $\frac{\partial L}{\partial \mathbf{F}_{*j}}$ using the chain rule. These derivative values are used to update the coefficient matrix $\mathbf{W}_x$, which is then fed into six layers CNN to update the parameters of $\varphi(x)$ in each layer via the BP algorithm.

Similar to the optimization of $\mathbf{W}_x$, we optimize $\mathbf{W}_y$ on the data modality $\mathbf{Y}$ with $\mathbf{W}_x$ and $\mathbf{B}$ fixed. The derivative values are similarly used to update the coefficient matrix $\mathbf{W}_y$, which is then fed into the adopted two-layer network to update the parameters of $\phi(y)$ in each layer via the BP algorithm.

\textbf{Optimize $\mathbf{B}$ with $\mathbf{W}_x$ and $\mathbf{W}_y$ fixed}:
When $\mathbf{W}_x$ and $\mathbf{W}_y$ are optimized and fixed, $\mathbf{F}$ and $\mathbf{G}$ are also determined, then the minimization problem in Eq. (\ref{eq4}) is equal to a maximization as follows:
\begin{flalign}
\label{eq12}
\begin{split}
\mathop {\max }\limits_{\mathbf{B}} tr(\lambda \mathbf{B}^T(\mathbf{F}\!+\!\mathbf{G}))\!=\!tr(\mathbf{B}^T\mathbf{U})\!=\!\sum_{i,j}\mathbf{B}_{ij}\mathbf{U}_{ij}
\end{split}
\end{flalign}
where $\mathbf{B}\in{\{-1,+1}\}^{n\times c}, \mathbf{U}={\lambda}(\mathbf{F}+\mathbf{G})$. It is easy to observe that the binary code $\mathbf{B}_{ij}$ should keep the
same sign as $\mathbf{U}_{ij}$. Therefore, we have:
\begin{equation}
\label{eq13}
\mathbf{B}=sign(\mathbf{U})=sign({\lambda}(\mathbf{F}+\mathbf{G}))
\end{equation}
%{\color{red}Currently we do differentiate the hash coding functions and the deep learning parameters. A differentiation is better, which can clearly show how we learn the parameter of DL and hashing functions.}

The whole procedure of RDCMH and entire iterative process for solving Eq. (\ref{eq4}) are summarized in Algorithm \ref{alg1}.
%We will empirically study the convergency of RDCMH in the next Experimental Section.
\begin{algorithm}[h!t]
\caption{RDCMH: Ranking based Deep Cross-Modal Hashing}
\label{alg1}
\begin{algorithmic}[1]
\REQUIRE Two modality data matrix $\mathbf{X}$ and $\mathbf{Y}$, and the corresponding label matrix $\mathbf{Z}$
\ENSURE Hashing coefficient matrices $\mathbf{W}_x$ and $\mathbf{W}_y$, the binary code matrix $\mathbf{B}$.
\STATE Initialize neural network parameters of $\varphi(x)$ and $\phi(y)$, mini-batch size
$n_x$ = $n_y$ = 128, and the number of iterations $iter$, $t=1$.
\STATE Calculate the similarity matrix $\mathbf{S}^{xx}$, $\mathbf{S}^{yy}$, $\mathbf{S}^{xy}$ and $\mathbf{S}^{yx}$.

\WHILE{$t<iter$ or not converged}
\STATE Randomly sample $n_x$ ($n_y$) triplets from $\mathbf{X}$ ($\mathbf{Y}$) to construct a mini-batch.
\STATE For each sampled triplet ($x_q$, $x_i$ , $x_j$) (or ($y_q$, $y_i$ , $y_j$)) in the mini-batch, compute $\mathbf{F}$ and $\mathbf{G}$ in Eq. (\ref{eq7}) by forward propagation;
\STATE Update coefficient matrix $\mathbf{W}_x$,$\mathbf{W}_y$ using Eqs. (\ref{eq9}-\ref{eq11});
\STATE Update the network parameters of $\varphi(x)$ ($\phi(y)$) based on $\mathbf{W}_x$ ($\mathbf{W}_y$) and back propagation;
\STATE Update $\mathbf{B}$ according to Eq. (\ref{eq13});
\STATE $t=t+1$.
\ENDWHILE
\end{algorithmic}
\end{algorithm}

\section{Experiment}

\subsection{Datasets}
We use three benchmark datasets: Nus-wide, Pascal VOC, and Mirflicker to  evaluate the performance of RDCMH. % The statistics of these  datasets are shown in Table \ref{Table1}.
Each dataset include two modalities (image and text), but RDCMH can also be applied to other data modalities. For $\geq 3$ modalities, we just need to compute the ranking lists for each modality and optimize it by minimizing the inconsistency of each ranking list between any pairwise modality.

\emph{Nus-wide}{\footnote{http://lms.comp.nus.edu.sg/research/NUS-WIDE.htm}} contains 260,648 web images,
and some images are associated with textual tags.
It is a multi-label dataset where each point is annotated
with one or several labels from 81 concept labels.
%We select 195,834 image-text pairs that belong to the 21 most
%frequent concepts.
The text for each point is represented as
a 1000-dimensional bag-of-words vector. The hand-crafted
feature for each image is a 500-dimensional bag-of-visual
words (BOVW) vector. %This dataset can be downloaded from %\href{http://lms.comp.nus.edu.sg/research/NUS-WIDE.htm}

\emph{Wiki}{\footnote{https://www.wikidata.org/wiki/Wikidata}} is generated from a group of 2866 Wikipedia documents. Each document is an image-text pair  labeled with 10 semantic classes. The images are represented by 128-dimensional SIFT feature vectors. The text articles are represented as probability distributions over 10 topics, which are derived from a Latent Dirichlet Allocation (LDA) model.% This dataset can be obtained from %\href{https://www.wikidata.org/wiki/Wikidata}

\emph{Mirflickr}{\footnote{http://press.liacs.nl/mirflickr/mirdownload.html}} originally contains 25,000 instances collected from Flicker. Each instance consists of an image and its associated textual tags, and is manually annotated with one or more labels,  from a total of 24  semantic labels. %To avoid noise, here we remove textual tags that appear less than 20 times in the dataset, and then exclude instances without textual tags or semantic %labels. This process gives us 16,738 instances.
The text for each point is represented as a 1386-dimensional bag-of-words vector. For the hand-crafted feature based method, each image is represented by a 512-dimensional GIST feature vector. %This dataset can be downloaded from %\href{http://press.liacs.nl/mirflickr/mirdownload.html}

%\begin{table}
%\label{Table1}
%\centering
%\caption{The statistics of Nus-wide, Mirflicker and Wiki  datasets.}
%\begin{tabular}{c|r|r|r|c}
%\hline
%Dataset & \#samples & \#labels & dimensions & modals\\\hline
%NUS-Wide & 260648 & 81 & 1000/500 & 2 \\
%Wiki & 2866 & 10 & 128/10 & 2 \\
%Mirflicker & 25000 & 24 & 500/150 &  2\\
%\hline
%\end{tabular}\\
%\label{Table1}
%\end{table}
%

\subsection{Evaluation metric and Comparing Methods}
We use the widely used Mean Average Precision (MAP)  to measure the retrieval performance of all cross-view hashing methods.
%The formal definition of MAP is  as follows:
%\begin{equation}
%MAP=\frac{1}{|\mathcal{Q}|}\sum\limits_{i=1}^{|\mathcal{Q}|}\frac{1}{t_i} {\sum\limits_{r=1}^{t_i}{P(r)\delta(r)}}
%\end{equation}
%where $\mathcal{Q}$ is the query set with  size equals to $|\mathcal{Q}|$. For the $i$-th query, $\frac{1}{t_i} {\sum\limits_{r=1}^{t_i}{P(r)\delta(r)}} $ denotes the average precision(AP), $t_i$ is the number of ground-truth relevant instances in the retrieval set, and $P(r)$ denotes the precision of the top $r$ retrieved results. $\delta(r)=1$ if the $r$-th retrieved result is a true neighbor of the query, otherwise $\delta(r)=0$.
A larger MAP value corresponds to a better retrieval performance.

Seven state-of-the-art and related cross-modal hashing methods are
used as baselines for comparison, including Cross-modal Similarity Sensitive Hashing (CMSSH) \cite{Bronstein2010Data}, Semantic Correlation Maximization (SCM-seq and SCM-orth) \cite{Zhang2014Large}, Semantics Preserving Hashing (SePH) \cite{Lin2017Cross}, Deep Cross-modal Hashing (DCMH) \cite{Jiang2017Deep}, Correlation Hashing Network (CHN) \cite{Cao2016Correlation} and Collective Deep Quantization(CDQ)\cite{Cao2017collective}. Source codes of these baselines are kindly provided by the authors and the input parameters of these baselines are specified according to the suggestion of the papers. As to RDCMH, we set the mini-batch size for gradient descent to 128, and set dropout rate as 0.5 on the fully connected layers to avoid overfitting. The regularization parameter $\lambda$ in Eq. (\ref{eq4}) is set to 1, and the number of iterations for optimizing Eq. (\ref{eq4}) is fixed to 500. We empirically found RDCMH generally converges in no more than 500 iterations on all these datasets. The length of the semi-supervised sematic ranking list used for training is set to 5. Namely, we divide the ranking list (i.e., ${\{r^{xy}_j}\}_{j=1}^n$ into 5 bins and randomly pick three points from three different bins to form a triplet for training. By doing so, we can not only capture different levels of semantic similarity, but also avoid optimizing too much triplets, whose maximum number is cubic to the number of samples. Our preliminary study shows that DRCMH holds relatively stable performance when the number of bins $\geq 4$.

\section{Results and Analysis}

\subsection{Search Accuracies}

\begin{table*}[h!t]
\label{Table2}
\centering
\scriptsize
\caption{
Results (MAP) on Mirflickr , Nus-wide and Wiki  dataset. }
\begin{tabular}{p{0.8cm}|p{1.3cm}|p{0.78cm}|p{0.78cm}|p{0.78cm}|p{0.78cm}||p{0.78cm}|p{0.78cm}|p{0.78cm}|p{0.78cm}||p{0.78cm}|p{0.78cm}|p{0.78cm}|p{0.78cm}}
\hline
 &  & \multicolumn{4}{c}{\textbf{Mirflickr}} & \multicolumn{4}{c}{\textbf{Nus-wide}} & \multicolumn{4}{c}{\textbf{Wiki}}\\\hline
 &Methods &16bits  &32bits &64bits &128bits &16bits  &32bits &64bits &128bits  &16bits  &32bits &64bits &128bits	\\\hline
 \multirow{8}{*}{\tabincell{c}{Image\\ vs.\\ Text}}
&  $\textbf{CMSSH}$ &	$0.5616$ &	$0.5555$ &	$0.5513$ &	$0.5484$   &	 $0.3414$ &	$0.3336$ &	$0.3282$ &	$0.3261$ & $0.1694$ &	$0.1523$ &	$0.1447$ &	$0.1434$   \\
&  $\textbf{SCM-seq}$ &$0.5721$ &	$0.5607$ &	$0.5535$ &	$0.5482$ &	$0.3623$ &	$0.3646$ &	$0.3703$ &	$0.3721$ & $0.1577$ &	$0.1434$ &	$0.1376$ &	$0.1358$   \\
&  $\textbf{SCM-orth}$ &	$0.6041$ &	$0.6112$ &	$0.6176$ &	$0.6232$  &	$0.4651$ &	$0.4714$ &	$0.4822$ &	$0.4851$ & $0.2341$ &	$0.2411$ &	$0.2443$ &	$0.2564$   \\
&  $\textbf{SePH}$ &	$0.6573$ &	$0.6603$ &	$0.6616$ &	$0.6637$   & $0.4787$ &	$0.4869$ &	$0.4888$ &	$0.4932$ & $0.2836$ &	$0.2859$ &	$0.2879$ &	$0.2863$  \\
&  $\textbf{DCMH}$ &$0.7411$ &	$0.7465$ &	$0.7485$ &	$0.7493$ & $0.5903$ &	$0.6031$ &	$0.6093$ &	$0.6124$  &  $0.2673$ &	$0.2684$ &	$0.2687$ &	$0.2748$ \\
&  $\textbf{CHN}$ &	$0.7438$ &	$0.7485$ &	$0.7511$ &	$0.7595$  &	 $0.6012$ &	$0.6028$ &	$0.6059$ &	$0.6121$ & $0.2534$ &	$0.2677$ &	$0.2681$ &	$0.2684$   \\
&  $\textbf{CDQ}$ & $0.7604$ &	$0.7631$ &	$0.7745$ &	$0.7738$ & $0.6203$ &	$0.6253$ &	$0.6274$ &	$0.6284$ & $0.2873$ &	$0.2834$ &	$0.2831$ &	$0.2901$ \\
\cline{2-14}
%&  $\textbf{DRCMH-NW}$ &	${\mathbf{0.1062}}$ &	${\mathbf{0.1064}}$  &	${\mathbf{0.1067}}$ &	${\mathbf{0.1058}}$\\
&  $\textbf{RDCMH}$ &	${\mathbf{0.7723}}$ &	${\mathbf{0.7735}}$ &		${\mathbf{0.7789}}$ & ${\mathbf{0.7810}}$&	${\mathbf{0.6231}}$ &		${\mathbf{0.6236}}$ &	${\mathbf{0.6273}}$ &${\mathbf{0.6302}}$ & ${\mathbf{0.2943}}$ &		${\mathbf{0.2968}}$ &	${\mathbf{0.3001}}$ &${\mathbf{0.3042}}$\\\hline\hline

  \multirow{8}{*}{\tabincell{c}{Text \\ vs. \\ Image}}
&  $\textbf{CMSSH}$ &	 $0.5616$ &	$0.5551$ &	$0.5506$ &	$0.5475$  &	 $0.3392$ &	$0.3321$ &	$0.3272$ &	$0.3256$ &  $0.1578$ &	$0.1384$ &	$0.1331$ &	$0.1256$ \\
&  $\textbf{SCM-seq}$ &	 $0.5694$ &	$0.5611$ &	$0.5544$ &	$0.5497$  &	$0.3412$ &	$0.3459$ &	$0.3472$ &	$0.3539$ & $0.1521$ & $0.1561$ &	$0.1371$ &	$0.1261$  \\
&  $\textbf{SCM-orth}$ &	$0.6055$ &	$0.6154$ &	$0.6238$ &	$0.6299$    &	$0.437$ &	$0.4428$ &	$0.4504$ &	$0.1235$  & $0.2257$ &	$0.2459$ &	$0.2482$ &	$0.2518$ \\
&  $\textbf{SePH}$ & $0.6481$ &	$0.6521$ &	$0.6545$ &	$0.6534$    & $0.4489$ &	$0.4539$ &	$0.4587$ &	$0.4621$ &  ${\mathbf{0.5345}}$ &	${\mathbf{0.5351}}$ &	${\mathbf{0.5471}}$ &	${\mathbf{0.5506}}$ \\
&  $\textbf{DCMH}$& $0.7827$ &	$0.7901$ &	$0.7932$ &	$0.7956$   & $0.6389$ &	$0.6511$ &	$0.6571$ &	$0.6589$ & $0.2712$ &	$0.2751$ &	$0.2812$ &	$0.2789$  \\
&  $\textbf{CHN}$ &	 $0.7402$ &	$0.7435$ &	$0.7463$ &	$0.7481$   &	$0.6415$ &	$0.6426$ &	$0.6435$ &	$0.6478$ &  $0.2416$ &	$0.2456$ &	$0.2483$ &	$0.2512$   \\
&  $\textbf{CDQ}$ & $0.7856$ &	$0.7841$ &	$0.7892$ &	$0.7931$ & $0.6531$ &	$0.6579$ &	$0.6613$ &	$0.6658$ & $0.2901$ &	$0.2847$ &	$0.3001$ &	$0.3021$ \\
\cline{2-14}
&  $\textbf{RDCMH} $ &	${\mathbf{0.7931}}$ &		${\mathbf{0.7924}}$ &	${\mathbf{0.8001}}$ &${\mathbf{0.8024}}$&	${\mathbf{0.6641}}$ &		${\mathbf{0.6685}}$ &	${\mathbf{0.6694}}$ &${\mathbf{0.6703}} $&	${{0.2931}}$ &		${{0.2956}}$ &	${{0.3012}}$ &${{0.3035}}$\\\hline

\end{tabular}
\label{Table2}
\end{table*}

\begin{table*}[h!t]
\label{Table3}
\centering
\scriptsize
\caption{
Results (MAP) on Mirflickr , Nus-wide and Wiki  dataset with 70\% unlabeled data. }
\begin{tabular}{p{0.8cm}|p{1.3cm}|p{0.78cm}|p{0.78cm}|p{0.78cm}|p{0.78cm}||p{0.78cm}|p{0.78cm}|p{0.78cm}|p{0.78cm}||p{0.78cm}|p{0.78cm}|p{0.78cm}|p{0.78cm}}
\hline
 &  & \multicolumn{4}{c}{\textbf{Mirflickr}} & \multicolumn{4}{c}{\textbf{Nus-wide}} & \multicolumn{4}{c}{\textbf{Wiki}}\\\hline
 &Methods &16bits  &32bits &64bits &128bits &16bits  &32bits &64bits &128bits  &16bits  &32bits &64bits &128bits	\\\hline
 \multirow{8}{*}{\tabincell{c}{Image\\ vs.\\ Text}}
&  $\textbf{CMSSH}$ &	$0.1384$ &	$0.1363$ &	$0.1332$ &	$0.1293$ &	 $0.0731$ &	$0.0723$ &	$0.0721$ &	$0.0716$ & $0.0146$ &	$0.0141$ &	$0.0126$ &	$0.0118$    \\
&  $\textbf{SCM-seq}$  & $0.1419$ &	$0.1391$ &	$0.1358$ &	$0.1331$ &	$0.0741$ &	$0.0734$ &	$0.0721$ &	$0.0711$ & $0.0162$ &	$0.0146$ &	$0.0148$ &	$0.0126$    \\
&  $\textbf{SCM-orth}$ &	$0.1321$ &	$0.1345$ &	$0.1386$ &	$0.1413$ &	$0.1056$ &	$0.1062$ &	$0.1074$ &	$0.1089$ &$0.0174$ &	$0.0158$ &	$0.0136$ &	$0.0109$    \\
&  $\textbf{SePH}$ &	$0.1511$ &	$0.1532$ &	$0.1541$ &	$0.1538$ & $0.1256$ &	$0.1249$ &	$0.1289$ &	$0.1291$ & $0.0674$ &	$0.0671$ &	$0.0684$ &	$0.0681$  \\
&  $\textbf{DCMH}$ & $0.1423$ &	$0.1435$ &	$0.1452$ &	$0.1468$ & $0.1055$ &	$0.1056$ &	$0.1059$ &	$0.1064$ &  $0.0541$ &	$0.0514$ &	$0.0553$ &	$0.0584$  \\
&  $\textbf{CHN}$ &	$0.1344$ &	$0.1357$ &	$0.1402$ &	$0.1431$ &	 $0.1125$ &	$0.1134$ &	$0.1151$ &	$0.1156$ & $0.0584$ &	$0.0602$ &	$0.0608$ &	$0.0611$   \\
&  $\textbf{CDQ}$ &	$0.1431$ &	$0.1423$ &	$0.1462$ &	$0.1433$ &	 $0.1242$ &	$0.1241$ &	$0.1195$ &	$0.1162$ & $0.0623$ &	$0.0637$ &	$0.0645$ &	$0.0681$   \\
\cline{2-14}
%&  $\textbf{DRCMH-NW}$ &	${\mathbf{0.1062}}$ &	${\mathbf{0.1064}}$  &	${\mathbf{0.1067}}$ &	${\mathbf{0.1058}}$\\
&  $\textbf{RDCMH}$ &	${\mathbf{0.1842}}$ &	${\mathbf{0.1861}}$ &		${\mathbf{0.1875}}$ & ${\mathbf{0.1889}}$&	${\mathbf{0.1634}}$ &		${\mathbf{0.1656}}$ &	${\mathbf{0.1674}}$ &${\mathbf{0.1705}}$ & ${\mathbf{0.1026}}$ &		${\mathbf{0.1058}}$ &	${\mathbf{0.1073}}$ &${\mathbf{0.1106}}$\\\hline\hline

  \multirow{8}{*}{\tabincell{c}{Text \\ vs. \\ Image}}
&  $\textbf{CMSSH}$ &	$0.1297$ &	$0.1343$ &	$0.1368$ &	$0.1392$ &	$0.0744$ &	$0.0751$ &	$0.0754$ &	$0.0758$ &  $0.0119$ &	$0.0119$ &	$0.0118$ &	$0.0117$  \\
&  $\textbf{SCM-seq}$ &	 $0.1331$ &	$0.1366$ &	$0.1395$ &	$0.1429$ &	$0.0756$ &	$0.0759$ &	$0.0765$ &	$0.0763$ & $0.0127$ &	$0.0122$ &	$0.0121$ &	$0.0118$   \\
&  $\textbf{SCM-orth}$ &	$0.1321$ &	$0.1376$ &	$0.1393$ &	$0.1414$ &	$0.0961$ &	$0.1008$ &	$0.1034$ &	$0.1047$ & $0.0105$ &	$0.0108$ &	$0.0113$ &	$0.0113$    \\
&  $\textbf{SePH}$ & $0.1434$ &	$0.1455$ &	$0.1462$ &	$0.1471$ & $0.1194$ &	$0.1204$ &	$0.1228$ &	$0.1264$ & $0.0651$ &	$0.0631$ &	$0.0635$ &	$0.0629$ \\
&  $\textbf{DCMH}$& $0.1284$ &	$0.1256$ &	$0.1284$ &	$0.1351$ & $0.1033$ &	$0.1065$ &	$0.1069$ &	$0.1075$ & $0.0564$ &	$0.0572$ &	$0.0585$ &	$0.0591$  \\
&  $\textbf{CHN}$ &	 $0.1384$ &	$0.1342$ &	$0.1384$ &	$0.1432$ &	$0.1134$ &	$0.1142$ &	$0.1148$ &	$0.1156$ & $0.0534$ &	$0.0542$ &	$0.0554$ &	$0.0561$   \\
&  $\textbf{CDQ}$ &	 $0.1324$ &	$0.1362$ &	$0.1358$ &	$0.1402$ &	$0.1131$ &	$0.1156$ &	$0.1163$ &	$0.1189$ & $0.0546$ &	$0.0584$ &	$0.0563$ &	$0.0573$   \\
\cline{2-14}
%&  $\textbf{DRCMH-NW} $ &	${\mathbf{0.1073}}$ &		${\mathbf{0.1078}}$ &	${\mathbf{0.1081}}$ &	${\mathbf{0.1069}}$\\
&  $\textbf{RDCMH} $ &	${\mathbf{0.1769}}$ &		${\mathbf{0.1786}}$ &	${\mathbf{0.1821}}$ &${\mathbf{0.1833}}$&	${\mathbf{0.1549}}$ &		${\mathbf{0.1569}}$ &	${\mathbf{0.1553}}$ &${\mathbf{0.1764}} $&	${\mathbf{0.1038}}$ &		${\mathbf{0.1045}}$ &	${\mathbf{0.1072}}$ &${\mathbf{0.1089}}$\\\hline

\end{tabular}
\label{Table3}
\end{table*}

The MAP results for RDCMH and other baselines with handcrafted features on MIRFLICKR, NUS-WIDE and Wiki datasets are reported in Table  \ref{Table2}. Here, `Image vs. Text' denotes the setting where the query is an image and the database is text, and `Text vs. Image' denotes the setting where the query is a text and the database is image.

From Table  \ref{Table2}, we have the following observations.

(1) RDCMH outperforms almost all the other baselines, which demonstrate the superiority of our method in cross-modal retrieval. This superiority is because RDCMH  integrates the semantic ranking information into deep cross-modal hashing to preserve  better semantic structure information, and jointly optimizes the triplet ranking loss and quantitative loss to obtain more compatible parameters of deep feature representations and of hashing functions. SePH achieves  better results for text to image retrieval on Wiki. That is possible because its adaptability of probability-based strategy on small datasets.

%the performance of DRCMH keeps increasing. This pattern indicates that it can utilize longer hash codes for better preserving the semantic affinities.
(2)   An unexpected observation is that the performance of CMSSH and SCM-Orth decreases as the length of hash codes  increase. This may be caused by the imbalance between bits in the hash codes learnt by singular value decomposition or eigenvalue decomposition, these two decompositions are adopted these two approaches.

(3)  Deep hashing methods (DCMH, CHN, CDQ and DRCMH) have an improved performance than the others. This proves that deep feature learned from raw data is more compatible for hashing learning than hand-crafted features in cross-modal retrieval. DRCMH still outperforms DCMH, CDQ and CHN. This observation corroborates  the superiority of ranking-based loss and the necessity of jointly learning deep feature presentations and hashing functions.

To further verify the effectiveness of RDCMH in semi-supervised situation, we randomly mask all the labels of 70\% training samples. All the comparing methods then use the remaining labels to learn hash functions. Table  \ref{Table3} reports the results under different hash bits on  three datasets. All these methods manifest sharply reduced MAP values. RDCMH have higher MAP values than all the other baselines, and also outperforms SePH on the Wiki dataset. RDCMH is less affected by the insufficient labels than other methods. For example, the average MAP value of the second best performer CHN is reduced by 81.9\%, and that of RDCMH is 70.2\%. This is because label integrity has a significant impact on the effectiveness of supervised hashing methods.  In practice,  the pairwise semantic similarity between labeled data is reduced to 9\% ($3/10\times3/10$) in this setting. As a result, RDCMH also has a sharply reduced performance. All these comparing methods ask for sufficient label information to guide the hashing code learning. Unfortunately, these comparing methods disregard unlabeled data, which contribute to more faithfully explore the structure of data and to reliable cross-modal hashing codes. This observation proves the effectiveness of the introduced semi-supervised semantic measure in leveraging unlabeled data to boost the hashing code learning.

We conducted additional experiments on multi-label datasets with 30\% missing labels by randomly masking the labels of training data. The recorded results show that RDCMH again outperforms the comparing methods. Specifically, the average MAP value of the second best performer (CDQ) is 4\% less than that of RDCMH. Due to space limitation, the results are not reported here. Overall, we can conclude that RDCMH is effective in weakly-supervised scenarios.
%\begin{table}
%\label{Table4}
%\centering
%\caption{The results on Mirflickr with 30\% missing labels.}
%\begin{tabular}{c||r|r|r|c}
%\hline
%CDQ(Text vs. Image)& $0.6231$ &	$0.6234$ &	$0.6257$ &	$0.6284$ \\\hline
%CDQ(Image vs. Text)& 	$0.6237$ &	$0.6134$ &	$0.6189$ &	$0.611$ \\\hline
%RDCMH(Text vs. Image) & $0.6779$ &	$0.6791$ &	$0.6803$ &	$0.6805$ 	\\\hline
%RDCMH(Image vs. Text) & $0.6804$ &	$0.6811$ &	$0.6832$ &	$0.6854$ \\\hline
%\end{tabular}\\
%\label{Table4}
%\end{table}

\subsection{Sensitivity to Parameters}

We further explore the sensitivity of the scalar parameter $\lambda$ in Eq. (\ref{eq4}), and report the results on Mirflickr and Wiki in Fig. \ref{Fig2}, where the code length fixed as 16 bits. We can see that RDCMH is slightly sensitive to $\lambda$ with $\lambda \in [10^{-3},10^3]$, and achieves the best performance when $\lambda=1$. Over-weighting or under-weighting the quantitative loss have a negative impact to the performance, but not so significant. In summary, an effective $\lambda$ can be easily selected for RDCMH.

\begin{figure}[h!tbp]
\centering
{\label{Fig2}\includegraphics[width=8.2cm,height=3.1cm]{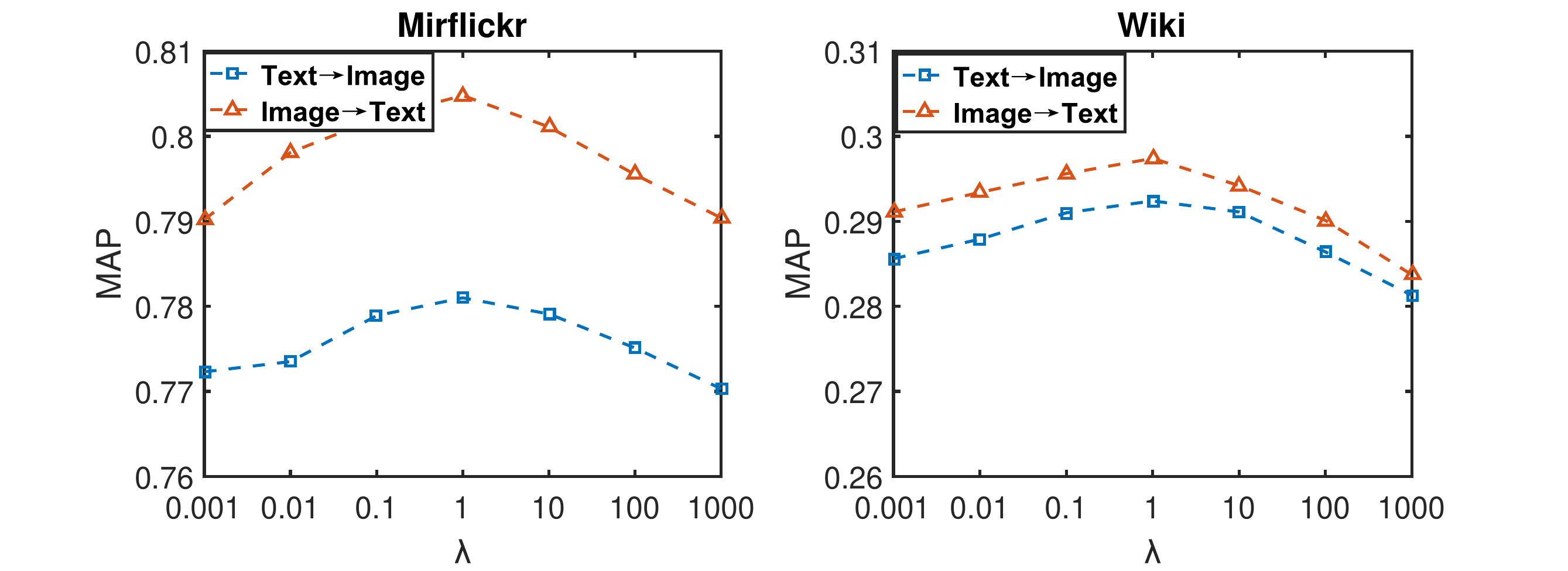}}
\caption{MAP vs. $\lambda$ on Mirfilcker and Wiki datasets.}
\label{Fig2}
\end{figure}

\begin{figure}[h!tbp]
\centering
{\label{Fig3}\includegraphics[width=8.2cm,height=3cm]{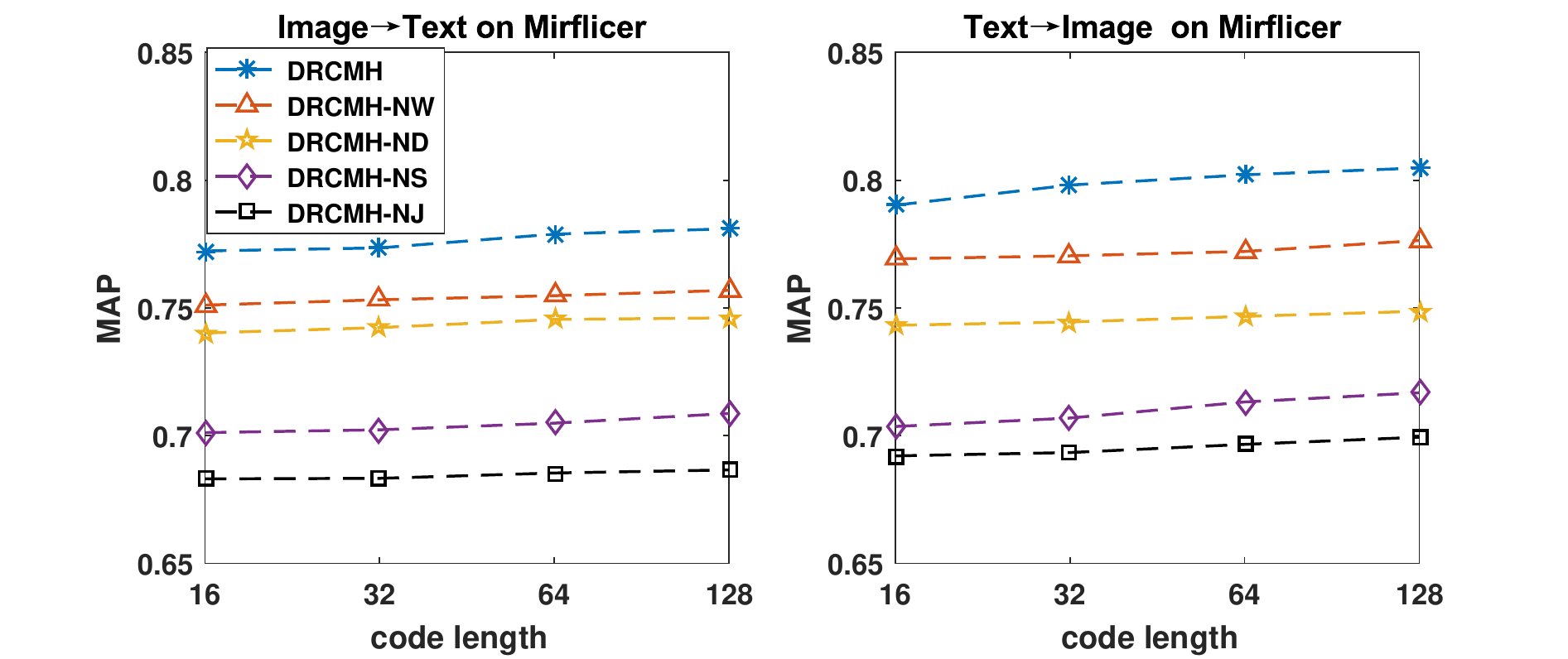}}
\caption{The results of different variants on Mirflicker.}
\label{Fig3}
\end{figure}
\subsection{Further Analysis}
To investigate the contribution components of RDCMH, we introduce four variants of RDCMH, namely RDCMH-NW, RDCMH-ND, RDCMH-NS and RDCMH-NJ. RDCMH-NW  disregards the weight $(1-s(i,j))$ and equally treats all the triplets; RDCMH-ND denotes the variant without deep feature learning, it directly uses the hand-crafted features to learn hashing functions during training.  RDCMH-NS simply obtains the ranking list by the number of shared labels, as done by \cite{Song2015Top,Zhao2015Deep}. RDCMH-NJ isolates deep feature learning and hashing functions learning, it first learns deep features and then generates hash codes based on the learnt features. Fig. \ref{Fig3} shows the results of these variants on the Mirfilcker dataset. The results on other datasets provide similar observations and conclusions, and are omitted here for space limit.

We can see RDCMH outperforms RDCMH-NW. This means the triplet ranking loss with adaptive weights can improve the cross-modal retrieval quality, since it assigns larger weights to more relevant points and smaller weights to the less relevant ones.  RDCMH also outperforms RDCMH-NS, which indicates that dividing the ranking lists into different levels based on the semi-supervised semantic similarity $\mathbf{S}$ is better than simply dividing by the number of shared labels,  which was adopted by  \cite{Zhao2015Deep,Lai2015Simultaneous}. Moreover, we can find that RDCMH achieves a higher accuracy than RDCMH-ND and RDCMH-NJ, which shows not only the superiority of  deep features than hand-crafted features in cross-modal retrieval, but also the advantage of simultaneous hash-code learning and deep feature learning.

\section{Conclusion}
In this paper, we proposed  a novel cross-modal hash function learning formwork (RDCMH) to seamlessly integrate deep feature learning with semantic ranking based hashing. RDCMH can preserve multi-level semantic similarity between multi-label objects for cross-modal hashing, and it also introduces a label and feature information induced semi-supervised semantic measure to leverage labeled and unlabeled data. Extensive experiments demonstrate that RDCMH outperforms other state-of-the-art hashing methods in cross-modal retrieval. The code of RDCMH is available at \url{mlda.swu.edu.cn/codes.php?name=RDCMH}.
%We will investigate probability-based strategy for cross-modal hashing with ranking information in our future work.

\section{Acknowledgments}

%We appreciate the reviewers for their insightful comments, which helped improving the paper.
This work is supported by NSFC (61872300, 61741217, 61873214, and 61871020), NSF of CQ CSTC (cstc2018jcyjAX0228, cstc2016jcyjA0351, and CSTC2016SHMSZX0824), the Open Research Project of Hubei Key Laboratory of Intelligent Geo-Information Processing (KLIGIP-2017A05), and the National Science and Technology Support Program (2015BAK41B04).

\bibliographystyle{aaai}
\bibliography{RDCMH_Bib}

\end{document}